\documentclass{article}

    \PassOptionsToPackage{round, compress}{natbib}

\usepackage[preprint]{neurips_2024}

\usepackage[utf8]{inputenc}              
\usepackage[T1]{fontenc}                 
\usepackage{hyperref}                    
\usepackage{url}                         
\usepackage{booktabs}                    
\usepackage{amsfonts}                    
\usepackage{nicefrac}                    
\usepackage{microtype}                   
\usepackage{xcolor}                      
\usepackage{multirow}                    
\usepackage{graphicx}                    
\usepackage[inkscapelatex=false]{svg}    
\usepackage{amsmath}                     

\usepackage{amsmath}
\usepackage{amsfonts}
\usepackage{amssymb}
\usepackage{wrapfig}
\usepackage{subcaption}
\usepackage{multirow}
 \usepackage{mathtools} 

\usepackage{verbatim}

\usepackage{anyfontsize}

\usepackage{microtype}
\usepackage{graphicx}
\usepackage{booktabs} 
\usepackage{multirow}
\usepackage{amsmath,amssymb}
\usepackage{booktabs}
\usepackage{caption,subcaption}

\usepackage{xcolor}
\definecolor{mygreen}{HTML}{3cb44b}
\definecolor{skyblue}{HTML}{beffff}
\definecolor{lightgreen}{HTML}{90ee90}

\usepackage{color, colortbl}

\definecolor{emerald}{rgb}{0.31, 0.78, 0.37}

\usepackage{tcolorbox}
\usepackage{enumitem}
\setitemize{itemsep=10pt,topsep=0pt,parsep=0pt,partopsep=0pt}
\usepackage{colortbl}

\usepackage{xcolor}
\definecolor{mygreen}{HTML}{3cb44b}
\colorlet{myyellow}{green!10!orange!90!}
\makeatletter

\usepackage{tikz}
\usetikzlibrary{arrows,shapes,automata,backgrounds,fit,petri}
\usepackage{adjustbox}

\newcommand{\RN}[1]{%
	\textup{\lowercase\expandafter{\it \romannumeral#1}}%
}
\usepackage{tabu}

\newcommand{\ie}[0]{\emph{i.e., }}

\newcommand{\eg}[0]{\emph{e.g., }}

\newcommand{\beq}{\vspace{0mm}\begin{equation}}
\newcommand{\eeq}{\vspace{0mm}\end{equation}}
\newcommand{\beqs}{\vspace{0mm}\begin{eqnarray}}
\newcommand{\eeqs}{\vspace{0mm}\end{eqnarray}}
\newcommand{\barr}{\begin{array}}
\newcommand{\earr}{\end{array}}

\usepackage{color, colortbl}
\definecolor{Gray}{gray}{0.93}

\usepackage{lipsum}

\usepackage{pifont}

\usepackage{makecell}

\usepackage{xcolor,amsmath}
\usepackage[linesnumbered,ruled,vlined]{algorithm2e}
\DontPrintSemicolon

\usepackage{xcolor}
\definecolor{mygreen}{HTML}{3cb44b}

\SetKwComment{Comment}{\color{green!50!black}\# }{}

\SetKwProg{Function}{def}{:}{}

\SetKwProg{For}{for}{:}{}
\SetKwProg{If}{if}{:}{}

\newcommand{\tablestyle}[2]{\setlength{\tabcolsep}{#1}\renewcommand{\arraystretch}{#2}\centering\footnotesize}

\hypersetup{
  colorlinks   = true, 
  urlcolor     = blue, 
  linkcolor    = blue, 
  citecolor   = violet 
}

\newtheorem{definition}{Definition}

\newcommand{\mypm}{\mathbin{\mathpalette\@mypm\relax}}
\newcommand{\@mypm}[2]{\ooalign{%
  \raisebox{.1\height}{$#1+$}\cr
  \smash{\raisebox{-.6\height}{$#1-$}}\cr}}
  
\newif\ifshowcomments
\showcommentstrue
\ifshowcomments
\newcommand{\mynote}[2]{\textcolor{blue}{\fbox{\bfseries\sffamily\scriptsize#1}}
  \textcolor{blue}{{$/*$\textsf{\emph{#2}}$*/$}}}
\ifshowcomments

\else
\newcommand{\mynote}[2]{}
\fi
\title{Differentially Private Fine-Tuning of Diffusion Models}

\author{%
  Yu-Lin~Tsai \thanks{National Yang Ming Chiao University}
  \And
  Yizhe~Li \thanks{Xi’an Jiaotong University}
  \And
  Zekai Chen \thanks{JPMorgan Chase \& Co.}
  \And
  Po-Yu Chen \thanks{Imperial College London}~~\footnotemark[3]
  \And
  Chia-Mu Yu \footnotemark[1]
  \And
  Xuebin Ren \footnotemark[2]
  \And
  Francois Buet-Golfouse \footnotemark[3]
}

\begin{document}

\maketitle
\begin{abstract}
  The integration of Differential Privacy (DP) with diffusion models (DMs) presents a promising yet challenging frontier, particularly due to the substantial memorization capabilities of DMs that pose significant privacy risks. 
Differential privacy offers a rigorous framework for safeguarding individual data points during model training, with Differential Privacy Stochastic Gradient Descent (DP-SGD) being a prominent implementation.
Diffusion method decomposes image generation into iterative steps, theoretically aligning well with DP's incremental noise addition. Despite the natural fit, the unique architecture of DMs necessitates tailored approaches to effectively balance privacy-utility trade-off. Recent developments in this field have highlighted the potential for generating high-quality synthetic data by pre-training on public data (\ie~ImageNet) and fine-tuning on private data, however, there is a pronounced gap in research on optimizing the trade-offs involved in DP settings, particularly concerning parameter efficiency and model scalability.
Our work addresses this by proposing a parameter-efficient fine-tuning strategy optimized for private diffusion models, which minimizes the number of trainable parameters to enhance the privacy-utility trade-off. We empirically demonstrate that our method achieves state-of-the-art performance in DP synthesis, significantly surpassing previous benchmarks on widely studied datasets (\eg with only 0.47M trainable parameters, achieving a more than 35\% improvement over the previous state-of-the-art with a small privacy budget on the CelebA-64 dataset). Anonymous codes available at \href{https://anonymous.4open.science/r/DP-LORA-F02F}{https://anonymous.4open.science/r/DP-LORA-F02F}.


\end{abstract}

\section{Introduction}
\label{sec:intro}

In the evolving landscape of generative AI, the burgeoning capabilities of models have raised profound concerns regarding data privacy~\citep{Wu2022MembershipIA,Duan2023AreDM}. Typical neural networks have been shown to inadvertently expose training data~\citep{Yin2021SeeTG,Carlini2023ExtractingTD}, prompting a surge in research aimed at enhancing privacy protections~\citep{Chen2021SyntheticDI} without significantly impairing model utility~\citep{Torfi2020DifferentiallyPS}. Among the notable advancements in this domain is the concept of \textit{Differential Privacy} (DP)~\citep{Dwork2006CalibratingNT,Dwork2014TheAF}, a rigorous framework designed to safeguard individual data points during the model training process. To implement DP in neural network training, \textit{Differential Privacy Stochastic Gradient Descent} (DP-SGD)~\citep{Abadi2016DeepLW} has been developed. This method involves modifying the traditional SGD process by clipping gradients and injecting noise, thus providing privacy guarantees for individual data samples used during training.

Recent emergence of generative models like \textit{diffusion models} (DMs)~\citep{Dhariwal2021DiffusionMB,Rombach2021HighResolutionIS,Ramesh2022HierarchicalTI,Balaji2022eDiffITD,Saharia2022PhotorealisticTD}, have demonstrated remarkable capabilities in synthesizing high-quality images and facilitating robust performance across various tasks. However, the extensive memorization capacity of these models has led to significant privacy concerns~\citep{Carlini2023ExtractingTD,Hu2023MembershipIO,Duan2023AreDM}, especially when trained on sensitive datasets~\citep{Ali2022SpotTF,Chambon2022RoentGenVF,Pinaya2022BrainIG}. This issue is exacerbated in domains where data cannot be freely shared or utilized due to ethical, legal, or privacy constraints.
As a pivotal shift from traditional generative adversarial networks (GANs)~\citep{Xie2018DifferentiallyPG,Torkzadehmahani2019DPCGANDP,Torfi2020DifferentiallyPS,harder2022pre}, DMs decompose the generation process into iterative steps. This feature theoretically makes DMs well-suited~\citep{Dockhorn2022DifferentiallyPD} for training under the constraints of DP, as the iterative nature of their training aligns with the incremental noise addition required by DP protocols.

Despite the theoretical compatibility of DMs with DP, the unique architecture of DMs, which relies on a gradual denoising process, requires tailored approaches~\citep{Dockhorn2022DifferentiallyPD,Ghalebikesabi2023DifferentiallyPD,Lyu2023DifferentiallyPL,Lin2023DifferentiallyPS} to effectively incorporate DP without overwhelming the model's capacity to learn from data. \citet{Dockhorn2022DifferentiallyPD} firstly suggest employing DP-SGD~\citep{Abadi2016DeepLW} for training diffusion models, though yield limited utility on datasets like CIFAR10 and CelebA. In response, \citet{Ghalebikesabi2023DifferentiallyPD} enhanced this approach by pretraining a large foundational generator on public data and then fine-tuning it with private data, achieving state-of-the-art results. Extremely recent work~\citep{Lyu2023DifferentiallyPL} extends vanilla diffusion scheme~\citep{Song2020ScoreBasedGM} to \emph{latent diffusion models} (LDMs)~\citep{Rombach2021HighResolutionIS}, demonstrating competitive results.
However, there is a pronounced gap in optimizing the \emph{privacy-utility trade-off} involved in different DP settings, \emph{particularly concerning parameter efficiency and model scalability}. Minimizing the number of trainable parameters can enhance the balance in DP because it reduces the amount of information the model needs to learn from the sensitive data, thus \emph{lowering the risk of privacy breaches}.
Recent research in non-private settings has introduced \textit{parameter-efficient fine-tuning} (PEFT) techniques (\ie LoRA by~\citet{Hu2021LoRALA}) to mitigate the issues related to storage and compute budgets~\citep{Dettmers2023QLoRAEF,Zhang2023AdaptiveBA}. 
Our work primarily aims to devise a strong and accurate parameter-efficient strategy with optimal privacy-utility trade-off by conducting a holistic study of different parameter-efficient settings under DP constraints.

Our primary contributions are: \textbf{1)} We demonstrate that our parameter-efficient fine-tuning methods achieve the state-of-the-art (SoTA) in DP image synthesis, significantly surpassing previous baselines on widely-studied benchmarks.
\textbf{2)} We thoroughly investigate and optimize parameter-efficient training settings under DP constraints and demonstrate that minimal trainable parameters can sufficiently yield competitive performance.
\textbf{3)} Our method facilitates a modular design where foundational pre-trained model can be quickly adapted for various downstream tasks with minimal modifications, enabling faster and more resource-efficient training of private diffusion models.
\section{Preliminary}
\label{sec:preliminary}

In this section, we provide a high-level overview to differential privacy and the latest work in relation to fine-tuning diffusion models (DMs) for incorporating differential privacy (DP).

\subsection{Differential Privacy}
\label{subsec:dp_fine_tune}

Differential privacy~\citep{Dwork2006CalibratingNT, Dwork2014TheAF} is the most popular approach to defend membership inference attack, where adversaries try to identify individuals or groups in the training data.

\begin{definition}
    A randomized mechanism $\mathcal{M}:\mathcal{D} \rightarrow \mathcal{R}$ satisfies $(\epsilon, \delta)$-differential privacy if for any two adjacent inputs $d,d' \in \mathcal{D}$, and any $S \subset \mathcal{R}$ fulfil the inequality below:
    \begin{equation}
    \mathbb{P}(\mathcal{M}(d) \in S) \leq e^\epsilon \mathbb{P}(\mathcal{M}(d') \in S) + \delta.
    \end{equation}
\end{definition}
where $\epsilon$ denote the privacy budget of which higher values indicate less privacy guarantee, whilst $\delta$ indicates the probability of information leak.

\subsection{Differentially Private Stochastic Gradient Descent}
\label{sec:dp-sgd}
Neural networks are commonly privatized using Differentially Private Stochastic Gradient Descent (DP-SGD)~\citep{Abadi2016DeepLW} or alternative DP optimizers like DP-Adam~\citep{McMahan2018AGA}. During each training iteration, the gradient for each mini-batch is clipped per example, and Gaussian noise is added. Formally, let \( l_i(f) := L(f, x_i, y_i) \) represent the learning objective with model parameters \( f \in \mathbb{R}^p \), input features \( x_i \), and label \( y_i \). The clipping function \( \texttt{clip}_C(v) : v \in \mathbb{R}^p \rightarrow \min(1, \frac{C}{\|v\|_2}) \cdot v \) ensures the input has a maximal \( \ell_2 \) norm of \( C \). For a minibatch \( B \) with \( |B| = B \) samples, the privatized gradient \( \hat{g} \) is given by \( \hat{g} = \frac{1}{B} \sum_{i \in B} \texttt{clip}_C(\nabla l_i(f)) + \frac{\sigma C}{B} \xi \), with \( \xi \sim \mathcal{N}(0, I_p) \) and \( I_p \in \mathbb{R}^{p \times p} \) being the identity matrix. The noise variance \( \sigma \), batch size \( B \), and training iterations are determined by the privacy budget \( (\epsilon, \delta) \). The choice of these hyperparameters significantly impacts model accuracy, making naive DP-SGD training challenging. Also, the DP guarantee can still be breached if an adversary makes a sufficient number of queries to a deferentially private model~\citep{dwork2008differential}. As part of this work, we use DP-SGD as our primary optimizer for all experiments and analysis.

\section{Parameter-efficient Differentially Private Latent Diffusion Models}
\label{sec:method}

In this section, we propose our solution, DP-LoRA. It implements a two-stage training process - 1) \textbf{pre-train} an LDM on a \textit{large} public dataset to ensure image generation quality, and then 2) \textbf{fine-tune} the LDM on a small \textit{private} dataset with limited privacy budgets (i.e., $\epsilon \leq M$) via Low-Rank Adapters (LoRA)~\citep{Hu2021LoRALA}.

\begin{figure}[]
\centering
\includegraphics[width=1.0\linewidth]{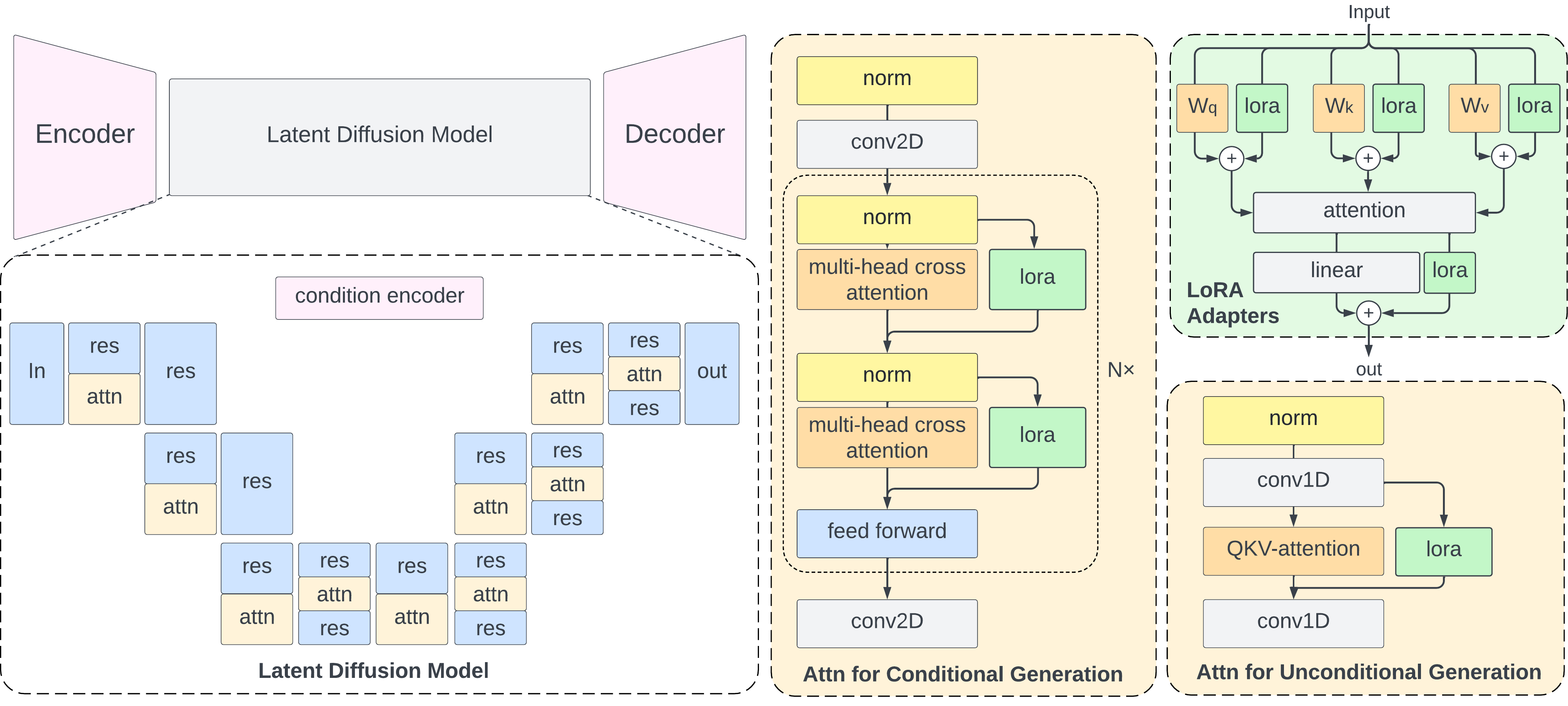}
\caption{An overview of our approach, DP-LoRA. After pre-training the autoencoder and the latent diffusion model (LDM), we fine-tune the pre-trained LDM by applying low-rank adaptors (LoRA) ~\citep{Hu2021LoRALA} to the attention blocks in LDM. In each attention block, we add LoRA not only to the QKV-attention matrices but also to the linear output projection layers for better performance.}
\label{fig:dp-lora_overview}
\end{figure}

\subsection{Fine-tuning LDM via Low-Rank Adaptation}
\label{subsec:solution_overview}   

Suppose $f(W_\texttt{PT};x)$ is a pre-trained model where $W_\texttt{PT}$ are the pre-trained parameters and $x$ denotes inputs. We can rewrite this function to incorporate additional trainable parameters $\theta$, where $\texttt{dim}(\theta) \ll \texttt{dim}(W_\texttt{PT})$, for fine-tuning. The new parameters are initialised to $\theta_0$ such that
\begin{equation}\label{eq:fine-tune_init}
    f_\texttt{FT}(W_\texttt{PT}, \theta_0; x) = f(W_\texttt{PT};x)
\end{equation}
From that, LoRA defines a fine-tuning method that is additive as
\begin{equation}\label{eq:fine-tune_lora}
    f_\texttt{FT}(W_\texttt{PT}, \theta; x) = f(W_\texttt{PT} + \lambda(\theta);x) 
\end{equation}
where the correction term $\lambda(\theta)$ is parameterised by $\theta$.
Here $W_\texttt{FT} = W_\texttt{PT}+\lambda(\theta)$ lies on a manifold passing through $W_\texttt{PT}$ of dimension $\texttt{dim}(\theta)$, which is much smaller than $\texttt{dim}(W_\texttt{PT})$.
Due to this observation, even if the parameters $\theta$ are very noisy due to the noise added during DP-SGD, the combined weights $W_\texttt{FT}$ remain in the manifold, ultimately preserving the image generation quality.

To incorporate DP by using the fine-tuning mechanism in Eq.\eqref{eq:fine-tune_lora}, we propose DP-LoRA. We also adopt the latent-diffusion models as they provide better generation qualities \cite{Lyu2023DifferentiallyPL} for images with higher resolutions. Figure \ref{fig:dp-lora_overview} illustrates an overview of our solution. The training mainly consists of two steps: 1) pre-training and 2) fine-tuning.

\paragraph{Pre-training.}
Initially, we train an auto-encoder with public data via SGD. The primary goal of this step is to reduce high-dimensional images into a lower-dimensional latent space~\citep{Rombach2021HighResolutionIS}. This transformation is crucial as it simplifies the subsequent training of the diffusion model by operating within this more manageable latent space. 
%
Subsequently, we train a LDM by tuning the entire model's parameters $W_{\texttt{PT}}$ without incorporating LoRA modules, using SGD. This step ensures that the LDM learns the underlying distribution of the data within the latent space, setting a robust foundation for high-quality image generation.


\paragraph{Fine-tuning.}
%
In the fine-tuning phase, we first convert private data into the lower-dimensional latent space using the pre-trained auto-encoder. We then fine-tune the LDMs using LoRA modules via DP-SGD combined with the \emph{noise multiplicity} proposed in DPDM \citep{Dockhorn2022DifferentiallyPD}. Adapters are specifically applied to the self-attention blocks responsible for learning unconditional image generations and the cross-attention blocks that handle conditional generation with text encoding, as illustrated in Figure~\ref{fig:dp-lora_overview}. Notably, the LoRA adapters are not only integrated into the Query, Key, and Value (QKV) attention modules within the attention blocks but also interact with the output projection layer, a linear feed-forward layer that restores the data to its original dimension. 

\subsection{Discussions}

\paragraph{Limitations of fully fine-tuning.}
In differentially private machine learning systems, there is a fundamental trade-off between the utility of the model and its privacy. In Section~\ref{sec:dp-sgd}, we clarify that for a given data sample $x_i$, the DP-SGD~\citep{Abadi2016DeepLW} limits the per-sample gradient $g(x_i)$ using the $\ell_{2}$ norm, meaning the \emph{clipping degree is proportional to the size of the network}. As a result, large networks experience significantly more disruption during full fine-tuning compared to their non-private counterparts, which substantially reduces their performance~\citep{Luo2021ScalableDP}. Second, diffusion models are much more computationally demanding to train because they work with higher-dimensional representations to generate entire images, rather than predicting logits~\citep{Ghalebikesabi2023DifferentiallyPD}. Inherently, DMs require more compute to converge in common settings (\eg \citet{Zagoruyko2016WideRN} trained Wide ResNet using $80k$ steps whereas~\citet{Ho2020DenoisingDP} trained DMs with $10\times$ more iterations). It remains challenging and crucial for optimal performance~\citep{De2022UnlockingHD} under high computational cost for fine-tuning DMs. Furthermore, the noise inherent in training diffusion models adds extra variance, which, when combined with the noise injected by DP-SGD, makes the training process even more challenging. Thus, efficiently and accurately training diffusion models with differential privacy is an urgent need.

\paragraph{Parameter-efficiency benefits private fine-tuning.}
In parallel to the increasing size of foundational models, \citet{Luo2021ScalableDP} argue that \emph{minimizing the number of trainable parameters is the key to improving the privacy-utility tradeoff of DP}. The intrinsic dimensionality hypothesis~\citep{Li2018MeasuringTI} also suggests that \emph{the minimum number of parameters required for training is much lower than the total model parameters}. Consistent observations have been found in the landscape of large language models (LLMs)~\citep{Aghajanyan2020IntrinsicDE}. Interestingly, as model size increases, the intrinsic dimension even decreases, leading to better zero-shot learning capability. We believe it offers an insightful perspective on the effectiveness of private parameter-efficient fine-tuning. Along this line, the concept of low-rank decomposition matrices~\citep{Hu2021LoRALA} significantly reduces the number of trainable parameters, making it feasible to adapt large models without requiring extensive computational resources (in Table~\ref{tb:unconditional_generation_fid_celeb64}, our method produces a FID of $8.4$ ($\epsilon=10$) on CelebA-64 with only $3.6\%$ trainable parameters, achieving an improvement beyond $50\%$ over previous SoTA). Importantly, any public large pre-trained model can be \emph{seamlessly augmented} with this lightweight modification for each individual downstream task benefiting from the modular design. The question then becomes:

\paragraph{Which modules are most worthwhile to optimize?} Trending work~\citep{Hertz2022PrompttoPromptIE,Zhang2023AddingCC} in DMs have demonstrated fine-tuning \emph{attention layers} for tasks like image editing and text-to-image generation is effective as attention mechanisms highlight important features in an image based on its context, which can vary across different data distributions. Following this spirit, \citet{Lyu2023DifferentiallyPL} suggested fine-tuning the attention modules and conditioning embedder that allows effective transfer of knowledge from public to private data distributions. Given that projection layers directly influence how features are transformed and integrated at different stages of the model, ensuring better alignment with the new data distribution, we also incorporate the out projection layer following attention module into fine-tuning.
Our ablation study in Table~\ref{tb:ablation_lora_components} also exemplifies the necessity fine-tuning both components (\eg FID of $7.71$ on CelebA-64 when fine-tuning both compared to a FID of $9.82$ whe excluding the projection layer). 
However, fine-tuning other parts, like \emph{ResBlocks}, can alter features significantly and reduce performance in private training~\citep{Lyu2023DifferentiallyPL}.








    

\section{Experiments}
\label{sec:exp}

\paragraph{Datasets.} 
To conduct experiment in DP settings, we utilized three image collections of varying complexity: the widely recognized MNIST~\citep{LeCun2005TheMD}, Fashion-MNIST~\citep{Xiao2017FashionMNISTAN}, the more intricate CelebA~\citep{liu2015faceattributes}, and CIFAR-10~\citep{Krizhevsky2009LearningML}. Additionally, we employed the high-resolution CelebA-HQ~\citep{Karras2017ProgressiveGO}, which consists of 256 × 256 images, to produce high-dimensional visuals. For class-conditional tasks, we used MNIST, CIFAR-10, Fashion-MNIST and CelebA-HQ. For unconditional tasks, we tested on CelebA with different resolutions. Regarding public datasets, we utilized EMNIST~\citep{Cohen2017EMNISTEM} as pretraining set corresponding to MNIST, CIFAR-10 for Fashion-MNIST, and rescaled ImageNet~\citep{Russakovsky2014ImageNetLS} for CIFAR-10, CelebA, and CelebA-HQ.

\paragraph{Current evaluation framework.}
To demonstrate the performance of DP-LoRA, we conducted a comprehensive analysis measures 1) image generation quality and 2) downstream utility.

The Fréchet Inception Distance (FID) \citep{heusel2017gans} is the predominant metric for evaluating the resemblance between synthetically produced and authentic images, and has therefore become a standard benchmark in the literature on DP image generation~\citep{Dockhorn2022DifferentiallyPD}. To quantitatively evaluate the quality of image generated, we adopt FID \citep{heusel2017gans} to measure the similarity between features distributions extracted from generated and real images.


We also adopt downstream classification accuracy on conditional generation tasks by training a classifier on the synthetic images of the same size as the real dataset and testing on the real test set. Regarding the downstream model architecture, we exploited a range of model architectures (that are popular in computer vision), including Convolutinoal Neural Network (CNN), Wide Residual Network (WRN) \citep{Zagoruyko2016WideRN} and ResNet-9 \citep{he2016deep}.

\paragraph{Baselines.}
To evaluate our approch, DP-LoRA, we compared it with other state-of-the-art baselines including DPDM~\citep{Dockhorn2022DifferentiallyPD}, 
DP-MEPF~\citep{harder2022pre}, DP-Diffusion~\citep{Ghalebikesabi2023DifferentiallyPD}, DP-LDMs~\citep{Lyu2023DifferentiallyPL}, PrivImage~\citep{li2024privimage} and dp-promise~\citep{wangdp}; see Section~\ref{sec:related} for more detailed introduction.

\paragraph{Implementations.} 
PyTorch~\citep{Paszke2019PyTorchAI} and Opacus~\citep{opacus} are used for for DP-SGD training and privacy accounting. Following the standard practice~\citep{Cao2021DontGM}, we set \(\delta = 10^{-5}\), for MNIST, Fashion-MNIST, CIFAR-10, and \(\delta = 10^{-6}\) for CelebA, ensuring \(\delta\) is smaller than the reciprocal of the number of training images. Nevertheless, our approach exhibits robust performance concerning hyperparameters, making them well-suited for privacy-critical applications; see App.~\ref{sec:appendix} for more details.

\begin{table}[!htb]
    \tablestyle{4pt}{1.2}
    \centering
    \begin{tabular}{l l c c c c c}
    \toprule
        Dataset                   & Method (Classifier) & $\epsilon=0.2$ & $\epsilon=1$   & $\epsilon=5$ & $\epsilon=10$ & $\epsilon=\infty$\\ 
        \midrule
        \rowcolor{Gray}
        \multirow{7}{*}{MNIST}    & \textbf{Ours(CNN)}     & -              & \textbf{96.4}            & -            & 97.9             & 98.35       \\
        \rowcolor{Gray}
                                  & \textbf{Ours(WRN)}     & -              & 94.8                     & -            & 97.8              & 98.16       \\ 
                                  & DP-LDM(CNN)            & -              & \underline{95.9$\mypm$0.1} & -            & 97.4$\mypm$0.1     & -           \\
                                  & DP-LDM(WRN)            & -              & -                        & -            & 97.5$\mypm$0.0     & -           \\
                                  & DPDM(CNN)              & -              & 95.2                     & -            & 98.1             & -           \\
                                  & DP-Diffusion(WRN)      & -              & -                        & -            & \textbf{98.6 }   & -           \\
                                  & dp-promise(CNN)        & -              & 95.8                     & -            & \underline{98.2} & -           \\
        \midrule
        \rowcolor{Gray}
        \multirow{8}{*}{CIFAR-10} & \textbf{Ours(ResNet9)} & -              & \textbf{67.76}   & \textbf{72.97}   & \textbf{73.98}   & 79.85       \\
        \rowcolor{Gray}
                                  & \textbf{Ours(CNN)}     & -              & 62.81 & 67.59 & 69.87 & 72.01       \\ 
                                  & DP-LDM(ResNet9)        & -              & 51.3$\mypm$0.1     & 59.1$\mypm$0.2     & 65.3$\mypm$0.3     & -           \\
                                  & DP-LDM(WRN)            & -              & -                & -                & 79.6$\mypm$0.3     & -           \\
                                  & DP-MEPF(ResNet9)       & -              & 28.9             & 47.9             & 48.9             & -           \\
                                  & DP-Diffusion(WRN)      & -              & -                & -                & 75.6             & -           \\
                                  & PrivImage+G(CNN)       & -              & 47.5             & 39.2             & 44.3             & -           \\
                                  & PrivImage+D(CNN)       & -              & \underline{66.2} & \underline{69.4} & \underline{68.8} & -           \\
        \midrule
        \rowcolor{Gray}
        \multirow{3}{*}{Fashion-MNIST} & \textbf{Ours(CNN)}& 62.9           & 67.8             & -                & 72.7           & -           \\
                                       & dp-promise(CNN)   & \underline{68.5}           & \textbf{81.6}             & -                & \underline{85.5}           & -           \\
                                       & DPDM(CNN)         & \textbf{72.3}           & \underline{79.4}             & -                & \textbf{86.2}           & -           \\ 
        \bottomrule
    \end{tabular}
    \vspace{2pt}
    \caption{Classification accuracy with class-conditional generations under different privacy levels (higher the better, best in \textbf{bold} and second best with \underline{underline}).}
    \vspace{-4mm}
    \label{tb:conditional_generation_cls}
\end{table}

\begin{table}[!ht]
\tablestyle{4pt}{1.2}
\centering
\begin{tabular}{l c c c}
\toprule
Class(gender) condition CelebA-HQ & $\epsilon=10$ & $\epsilon=5$ & $\epsilon=1$ \\
        \midrule
        \rowcolor{Gray}
        \textbf{Ours} & \textbf{17.2} & \textbf{18.2} & \textbf{20.0} \\
        DP-LDM & 19.0$\mypm$0.0 & 20.5$\mypm$0.1 & 25.6$\mypm$0.1 \\
        DP-MEPF  &200.8 & - & 293.3 \\
        \bottomrule
    \end{tabular}
    \vspace{2pt}
    \caption{FID result \emph{w.r.t} gender-conditional generations on CelebA-HQ (rank is set to 8, number of noise multiplicity is 4 with 10,000 samples).}
    \vspace{-5mm}
    \label{tb:conditional_generation_fid}
\end{table}

%
\subsection{Conditional Generations}
\label{subsec:conditional_generations}
%

\paragraph{Classification results.} Our experimental results, as summarized in Table~\ref{tb:conditional_generation_cls}, provide a comprehensive evaluation of the performance of various DPDM methods across multiple datasets and privacy levels, measured by the epsilon parameter ($\epsilon$). The datasets used, MNIST, CIFAR-10, and Fashion-MNIST, vary in complexity and are benchmarks commonly used to assess image classification models.
In the \textbf{MNIST} dataset, our CNN-based classifier achieves a notable accuracy of $96.4\%$ at $\epsilon=1$ and peaks at $97.9\%$ for $\epsilon=10$, indicating robust performance under stringent privacy constraints. The Wide Residual Network (WRN) variant of our model also performs competitively. Notably, the DP-Diffusion~\citep{Ghalebikesabi2023DifferentiallyPD} model with WRN achieves the highest accuracy of $98.6\%$ at $\epsilon=10$, suggesting that more trainable parameters are  likely deserved when private budget is high. For the \textbf{CIFAR-10} dataset, our method with ResNet9 stands out, consistently outperforming other models (with same classifier) across different $\epsilon$ values, with a peak accuracy of $73.98\%$ at $\epsilon=10$. This is significantly higher than the next best model DP-LDM~\citep{Lyu2023DifferentiallyPL} using the same backbone, which achieves a maximum of $65.3\%$ under similar conditions. The results underscore the effectiveness of our approach, particularly with complex image data where maintaining high utility under DP constraints is challenging.
\textbf{Fashion-MNIST} results further reinforce the capability of our CNN model, achieving a respectable $72.7\%$ accuracy at $\epsilon=10$. However, it is outperformed by both dp-promise~\citep{wangdp} and DPDM~\citep{Dockhorn2022DifferentiallyPD} models. The reason is that we use CIFAR-10 for pretraining (because LDM requires an auto-encoder), and converting CIFAR-10 to grayscale for training leads to poor auto-encoder performance. Alternatively, using EMNIST for training the auto-encoder results in outputs that are too line-based, leading to unsuitable results. In contrast, vanilla DMs (\eg DPDM and dp-promise) do not encounter this issue.
Overall, the results across different datasets and privacy levels illustrate the trade-offs between model utility and privacy. While our models generally perform competitively, especially in more complex datasets like CIFAR-10, there is variability in performance across different architectures and configurations. 

\paragraph{Generation on high-quality images.}
Table~\ref{tb:conditional_generation_fid} presents the FID scores for gender-conditional generations on the CelebA-HQ dataset at different privacy budgets (the CelebA-HQ is divided into two groups: male and female). Our method significantly outperforms the other models across all privacy levels, achieving FID scores of $17.2$, $18.2$, and $20.0$ for \(\epsilon=10\), \(\epsilon=5\), and \(\epsilon=1\) respectively. The improvement is more than $20\%$ when private budge is low, compared to the second best DP-LDM. In contrast, DP-MEPF exhibits significantly higher FID scores, with DP-LDM ranging from $19.0$ to $25.6$. These results demonstrate the superior performance and robustness of our approach in generating high-quality images under differential privacy constraints.

\begin{figure}[]
\centering
\includegraphics[width=0.24\linewidth]{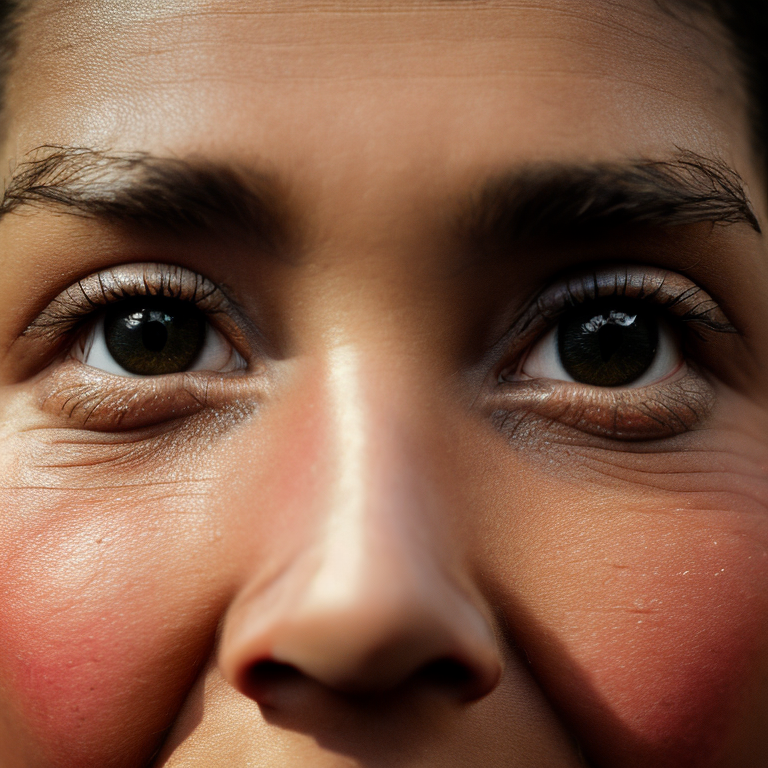}
\includegraphics[width=0.24\linewidth]{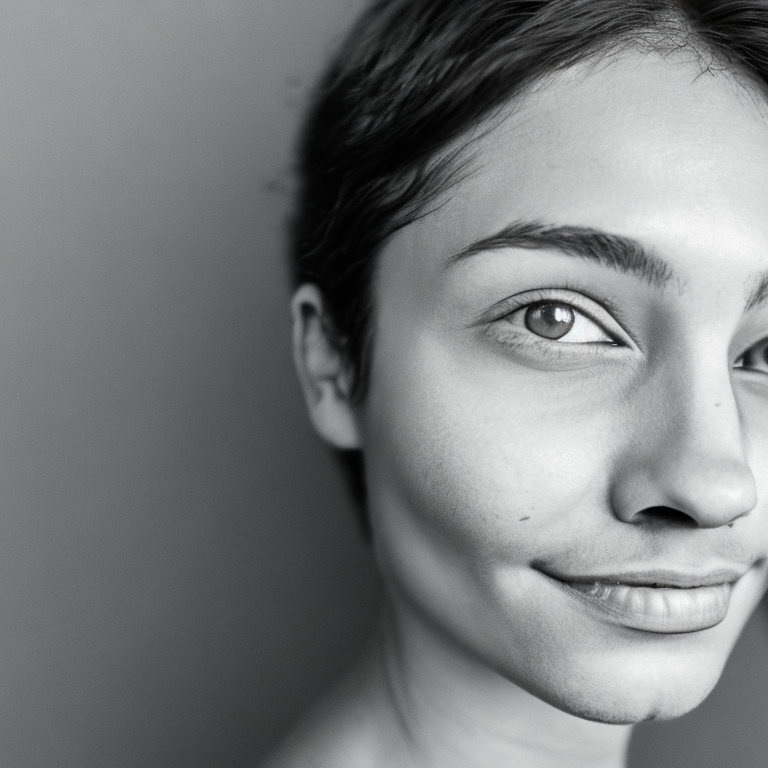}
\includegraphics[width=0.24\linewidth]{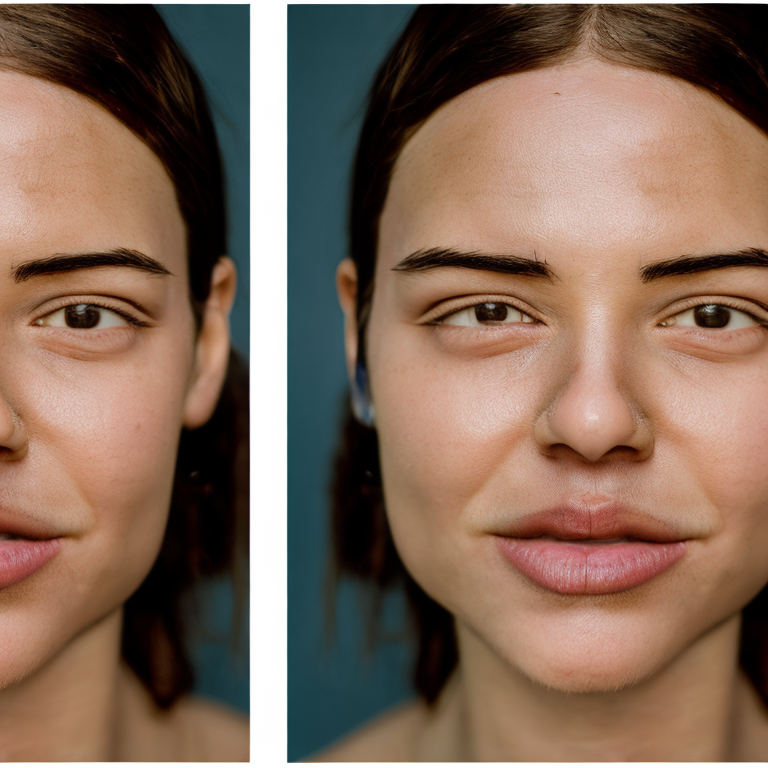}
\includegraphics[width=0.24\linewidth]{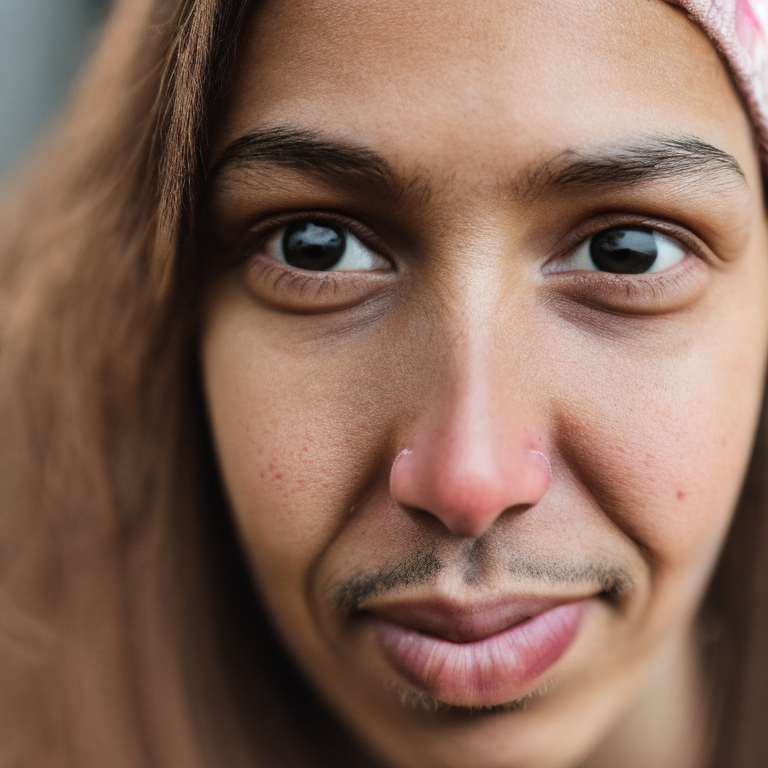}

\includegraphics[width=0.24\linewidth]{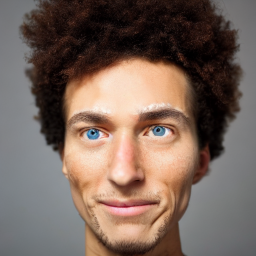}
\includegraphics[width=0.24\linewidth]{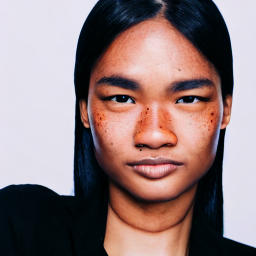}
\includegraphics[width=0.24\linewidth]{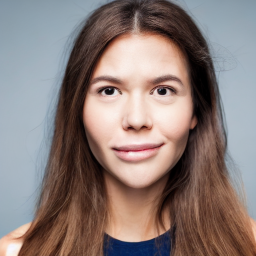}
\includegraphics[width=0.24\linewidth]{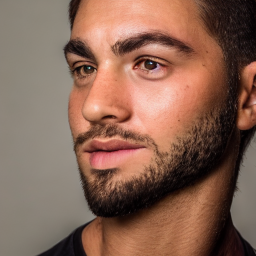}
\caption{Examples generated with the prompt: "\texttt{a good and full photo of <HF>}" with stable-diffusion-v1.5 as the foundational model.}
\vspace{-5mm}
\label{fig:example}
\end{figure}


\subsection{Unconditional Generations}
\label{subsec:unconditional_generations}

%
\begin{wraptable}{l}{6cm}
    \tablestyle{4pt}{1.2}
    \centering
    \vspace{-10pt}
    \begin{tabular}{l ccc}
    \toprule
        \multirow{2}{*}{Method}
                                 &\multicolumn{3}{c}{CelebA-64} \\
                                 & $\epsilon=1$ & $\epsilon=5$ & $\epsilon=10$ \\ 
        \midrule
        \rowcolor{Gray}
        \textbf{Ours}            & \textbf{12.0}    & \textbf{9.5}     & \textbf{8.4}    \\
        DP-LDMs                  & 21.1             & \underline{16.1}             & \underline{14.3}              \\
        DP-MEPF($\phi_1,\phi_2$) & 19.0             & 19.1             & 18.5              \\
        DP-MEPF($\phi_1$)        & \underline{18.4} & 16.5 & 17.4 \\
        PPRIVIMAGE+G             & 45.1             & 45.2             & 38.2             \\
        PPRIVIMAGE+D             & 71.4             & 52.9             & 49.3            \\
        dp-promise               & 29.1             & 26.2             & 25.3            \\
        \bottomrule
    \end{tabular}
    \vspace{1pt}
    \caption{Unconditioned generations results on CelebA-64 under different privacy levels (lower the better). Our method significantly outperforms other SoTAs.}
    \vspace{-10pt}
    \label{tb:unconditional_generation_fid_celeb64}
\end{wraptable}

The results presented in Table~\ref{tb:unconditional_generation_fid_celeb64} provide a detailed comparison of various algorithms' performance on the CelebA-64 dataset in terms of FID. See Figure~\ref{fig:example} for sample generations from CelebA-64 using Stable-Diffusion-v1.5 as the foundation.
Our algorithm consistently outperforms all other methods across all tested privacy levels ($\epsilon=1, 5, 10$) on the CelebA-64 dataset. Specifically, our model achieves the lowest FID scores of 12.0, 9.5, and 8.4 for $\epsilon=1, 5, \text{and} 10$, respectively. This indicates not only the effectiveness of our approach in generating high-quality images but also its robustness in maintaining performance even as privacy constraints are relaxed (higher $\epsilon$ values).
The DP-LDMs~\citep{Lyu2023DifferentiallyPL} show moderate performance with FID scores decreasing from 21.1 at $\epsilon=1$ to 14.3 at $\epsilon=10$. This trend suggests some improvement in image quality with reduced privacy constraints but still lags significantly behind our model. The DP-MEPF~\citep{harder2022pre} variants exhibit mixed results; while DP-MEPF($\phi_1$) shows a slight improvement over DP-MEPF($\phi_1,\phi_2$), both configurations perform worse than our model and DP-LDMs, indicating potential limitations of compressing latent embeddings to the average space (similar to textual-inversion~\citep{Gal2022AnII}) that may result in \emph{insufficient parameters for fine-tuning}.

\begin{wraptable}{r}{5cm}
    \tablestyle{3pt}{1.1}
    \centering
    \small
    \vspace{-10pt}
    \begin{tabular}{l cccc}
    \toprule
     & $\epsilon=1$ & $\epsilon=5$ & $\epsilon=10$ & $\epsilon=\infty$ \\
    \midrule
           ours       & 72.6 & 64.9 & 48.9 & 41.1\\
    \bottomrule
    \end{tabular}
    \vspace{1pt}
    \caption{We leverage textual-inversion and Stable-Diffusion-v1.5 to generate high-quality synthetic images from the MM-CelebA-HQ dataset, including 30,000 512$\times$512 samples.}
    \vspace{-10pt}
    \label{tb:textual-inversion}
\end{wraptable}

The dp-promise~\citep{wangdp} does not match the performance of our model or the DP-LDMs, with FID scores decreasing from 29.1 at $\epsilon=1$ to 25.3 at $\epsilon=10$. This is likely due to dp-promise still uses vanilla diffusion model which trains from a higher pixel space instead of latent space. Compared to LDM, this paradigm has \emph{more parameters to train and generally requires more compute to converge} which makes the optimization more challenging. Complete results in App.~\ref{sec:complete_results}.
Overall, the results underscore the superiority of our algorithm in generating high-quality images under various levels of differential privacy. The consistent outperformance across different $\epsilon$ values highlights the effectiveness of our approach in managing the trade-off between privacy and image quality. 

\subsection{Textual-Inversion}
Textual inversion~\citep{Gal2022AnII} is another popular method to enhance generation performance, by learning semantic embeddings which inverts into the text prompts space of a pre-trained model to generate semantically relevant images based on these embeddings. We also demostrate the results in Table~\ref{tb:textual-inversion} (despite the presence of NSFW content, which was not manually filtered, the FID score calculation includes these outputs to fairly assess the model's performance). Notably, we did not observe competitive results by leveraging this technique compared to the standard practice. We argue that \emph{generic textual embeddings struggle with accurately representing complex or highly detailed images}, especially for highly diverse datasets, potentially reducing the quality of the outputs.

\subsection{Ablation Study}
\label{subsec:ablation_study}

In this section, we delve into ablation studies designed to dissect the influence of various factors on the performance and efficiency of our model. Specifically, we examine the impact of choices in Low-Rank Adaptation (LoRA) parameters and ranks ($r$), steps of noise multiplicity ($k$), and the trade-offs involved in textual inversion partitioning and privacy. These studies aim to pinpoint critical elements that significantly affect the model's effectiveness and operational cost, providing insights into optimal configurations for balancing performance with private budget.

\begin{table}[]
    \tablestyle{3pt}{1.1}
    \centering
    \begin{tabular}{l cccc cccc}
    \toprule
    \multirow{2}{*}{Metric}
                              & \multicolumn{4}{c}{CelebA-32}
                              & \multicolumn{4}{c}{CelebA-64} \\
                                \cmidrule(r){2-5} \cmidrule(r){6-9}
                              & $k=1$      & $k=2$      & $k=4$       & $k=8$  & $k=1$      & $k=2$      & $k=4$       & $k=8$ \\
    \midrule
    FID  & 13.75  & 10.91 & 7.71 & \textbf{7.32} & 12.29 & 11.03 & \textbf{8.41} & 8.94  \\
    training time/epoch & 6~mins        & 9~mins        & 15~mins       & 28~mins  & 22~mins & 37~mins & 70~mins & 140~mins   \\ 
    \bottomrule
    \end{tabular}
    \vspace{2pt}
    \caption{Ablation study with respect the number of noise multiplicity steps ($k$) when fine-tuning on CelebA-32 and CelebA-64.}
    \vspace{-5mm}
    \label{tb:ablation_dpdm}
\end{table}

\paragraph{Number of noise multiplicity steps.}
Table~\ref{tb:ablation_dpdm} examines the impact of varying the number of noise multiplicity steps ($k$) on the FID score and training time per epoch for fine-tuning on CelebA-32 and CelebA-64 datasets. In general, as $k$ increases, the FID scores improve, indicating better image quality, with the lowest FID observed at $k=8$ for CelebA-32 ($7.32$) and $k=4$ for CelebA-64 with FID of $8.41$). However, this improvement comes at the cost of significantly increased training time, rising from 6 minutes to 28 minutes per epoch for CelebA-32 and from 22 minutes to 140 minutes per epoch for CelebA-64. In practice, we recommend setting $k$ in between 4 to 8 to obtain the optimal results given the the trade-off between image quality and computational efficiency.

\paragraph{Lower rank sufficiently yield competitive results.}

\begin{wraptable}{l}{5.5cm}
    \tablestyle{6pt}{1.2}
    \centering
    \vspace{-10pt}
    \begin{tabular}{l c c}
    \toprule
    Rank & FID & $\Delta$ \#Params \\
    \midrule
           $r=8$       & 8.09 / 10.01         & 359K / 239K \\
    \rowcolor{Gray}
           $r=16$      & \textbf{7.71 / 8.41} & 718K / 479K \\ 
           $r=32$      & 7.83 / 9.14          & 1.4M / 958K \\
           $r=64$      & 10.16 / 9.03         & 2.9M / 1.9M \\
    \bottomrule
    \end{tabular}
    \vspace{1pt}
    \caption{Ablation study with respect to different adapter ranks on both CelebA-32 (left) and CelebA-64 (right). Both results show consistent trend.}
    \vspace{-10pt}
    \label{tb:ablation_lora_rank}
\end{wraptable}

Table~\ref{tb:ablation_lora_rank} reveals a clear relationship between rank and generation performance. Lower ranks (\eg, $r=16$) achieve the best FID score of $7.71$ and $8.41$, indicating optimal balance between model complexity and utility. Increasing the rank to $32$ slightly degrades the FID to 7.83, suggesting diminishing returns with added complexity. Notably, the highest rank, $r=64$ with 2.9M additional parameters, results in a significantly worse FID score of $10.16$, indicating inefficiencies and diminished returns. This highlights \emph{a threshold beyond which increasing the rank is counterproductive}. Conversely, the lowest rank, $r=8$, achieves an FID of $8.09$, showing that while fewer parameters can still yield competitive results, moderate complexity ($r=16$) provides the best overall performance. This indicates that an optimal rank exists, balancing model capacity and generalization, which is consistent with the observation of~\citet{Luo2021ScalableDP}.

\paragraph{How do different components impact the performance?}

The results (see Table~\ref{tb:ablation_lora_components}) highlight the essential roles of the QKV matrices and the projection layer within the attention module for fine-tuning DP diffusion models. Including both components yields the best performance (FID of $7.71$ with mild additional parameters), as the QKV matrices are crucial for computing attention scores that capture complex dependencies, while the projection layer integrates these scores into the model's representations. Excluding the QKV matrices significantly worsens performance (FID of $11.78$), and omitting the projection layer also degrades results (FID of $9.82$), demonstrating that both are vital for maintaining image quality and optimizing parameter efficiency.

\begin{wraptable}{r}{6.5cm}
    \tablestyle{5pt}{1.2}
    \centering
    \vspace{-10pt}
    \begin{tabular}{l c c}
    \toprule
    FT Modules & FID & $\Delta$ \#Params \\
    \midrule
    \rowcolor{Gray}
            + QKV \& Project     & \textbf{7.71} / \textbf{8.41}   & 718K / 479K \\
            - QKV                & 11.78 / 12.89 & 249K / 159K \\
            - Project Layer      & 9.82 / 11.24  & 479K / 319K \\
    \bottomrule
    \end{tabular}
    \vspace{1pt}
    \caption{Ablation study with respect to different fine-tuning components on both CelebA-32 (left) and CelebA-64 (right). Two results show consistent trend.}
    \vspace{-10pt}
    \label{tb:ablation_lora_components}
\end{wraptable}

\section{Related Work}
\label{sec:related}

\paragraph{Diffusion models.}

Diffusion models have recently gained significant attention due to their robust performance in generating high-quality synthetic data. These models work by iteratively denoising a variable initially sampled from a simple distribution~\citep{Ho2020DenoisingDP,Song2020DenoisingDI,Dhariwal2021DiffusionMB}, gradually transforming it to match the target data distribution. Recent efforts have focused on enhancing the efficiency~\citep{Nichol2021GLIDETP,Nichol2021ImprovedDD,Saharia2022PhotorealisticTD} and scalability~\citep{Rombach2021HighResolutionIS,Peebles2022ScalableDM} of diffusion models, including techniques for faster sampling, latent representations and improved training stability. 

\paragraph{Differentially private image generation.}
Additionally, the application of differentially private mechanisms to diffusion models is an emerging research area, aiming to generate high-fidelity synthetic data while preserving individual privacy~\citep{Dockhorn2022DifferentiallyPD}. This line of work is crucial for advancing privacy-preserving machine learning and enabling the use of synthetic data in sensitive domains. A significant portion of studies has been dedicated to applying DP-SGD~\citep{Abadi2016DeepLW} to generative adversarial networks~\citep{Xie2018DifferentiallyPG,Torkzadehmahani2019DPCGANDP,Chen2020GSWGANAG} and variational autoencoders~\citep{Pfitzner2022DPDfVAESD}. With the rise of diffusion models~\citep{Rombach2021HighResolutionIS}, much of the recent research has shifted focus to applying DP-SGD to these models. Among which, ~\citet{Dockhorn2022DifferentiallyPD} first examined the use of DP-SGD in the context of diffusion models, which we also compare in Table~\ref{tb:conditional_generation_cls} and~\ref{tb:unconditional_generation_fid_celeb64}. \citet{Ghalebikesabi2023DifferentiallyPD} further scale up the diffusion models and discover that pretraining on public data followed by fine-tuning on private data is highly effective, achieving state-of-the-art results. DP-LDM~\citep{Lyu2023DifferentiallyPL} is introduced by using latent diffusion model defined on the lower-dimensional latent space has a significantly lower
number of parameters to fine-tune than the diffusion model defined on the pixel space. 
Meanwhile, other works have proposed custom architectures~\citep{Harder2020DPMERFDP,harder2022pre,chen2022dpgen,li2024privimage,wangdp}. DP-MEPF~\citet{harder2022pre}, for instance, pre-train a perceptual feature extractor using public data, then privatize the mean of the feature embeddings of the sensitive data records, and use the privatised mean to train a generative model. 
DPGEN \citep{chen2022dpgen} employ energy-based generative models trained on
differentially private scores, which are constructed by randomised responses. 
PRIVIMAGE \citep{li2024privimage} establishes a semantic query function using a public dataset.
%
%
DP-Promise \citep{wangdp} argue that we can apply DP noise to the first $S$ steps in the forward process, which promotes the model utility by reducing the injected noise.
However, given the gigantic size of modern DMs, the fine-tuning can still be expensive, and consequently, limit the utility of fine-tuned DMs.

\section{Conclusion}
\label{sec:conclusion}
In this paper, we explored the integration of Differential Privacy (DP) with diffusion models (DMs), addressing the substantial privacy risks posed by the memorization capabilities of these models. Our study focused on optimizing the privacy-utility trade-off through a parameter-efficient fine-tuning strategy that minimizes the number of trainable parameters, thus enhancing the model's privacy while maintaining high utility. We empirically demonstrated that our approach achieves state-of-the-art performance in DP synthesis, significantly surpassing previous benchmarks with a small privacy budget. This work highlights the potential of parameter-efficient techniques in advancing privacy-preserving generative models, paving the way for more scalable and practical applications in sensitive data domains. 

\bibliography{neurips_2024.bib}
\bibliographystyle{abbrvnat}

\newpage
\appendix
\section{Limitations}
\label{sec:limitations}
While our approach demonstrates significant advancements in the privacy-utility trade-off for diffusion models, several limitations remain. First, the computational cost of pre-training and fine-tuning, especially with large models, is substantial, necessitating high-end hardware like RTX 4090 and RTX 3090 GPUs. Second, our method assumes the availability of large public datasets for pre-training, which might not always be feasible. Additionally, although we mitigate privacy risks through DP-SGD, the inherent noise addition can still degrade model performance, particularly when dealing with very sensitive or highly variable data. Lastly, the manual integration of LoRA modules into specific components of the model introduces complexity (thought it's straightforward), it requires extensive tuning and experimentation to apply to other architectures or domains.

\section{Broader Impact}
\label{sec:impact}
The integration of differential privacy with diffusion models represents a critical step forward in ensuring data privacy while maintaining the utility of generative AI systems. This work addresses growing concerns about data security in an era of increasingly powerful AI models capable of memorizing and potentially exposing sensitive information. By demonstrating effective methods for balancing privacy and performance, our research contributes to the responsible deployment of AI technologies, particularly in sensitive fields such as healthcare, finance, and personal data applications. Furthermore, the parameter-efficient fine-tuning strategy offers a scalable solution that can be adapted to various models and tasks, promoting broader adoption of privacy-preserving techniques in the AI community. However, as with any technology, the potential for misuse remains; ensuring ethical guidelines and robust oversight will be essential as these methods are integrated into real-world applications.

\section{Implementation Details}
\label{sec:appendix}

All experimental models use PyTorch. Differential Privacy (DP) is implemented using Opacus\footnote{https://github.com/pytorch/opacus}. Parameter-Efficient Fine-Tuning (PEFT) and LoRA utilize the PEFT library provided by Huggingface\footnote{https://github.com/huggingface/peft}, with added support for Conv1D. Textual inversion is implemented using Huggingface's Diffusers library\footnote{https://github.com/huggingface/diffusers}. The overall codebase is based on the Latent Diffusion paper\footnote{https://github.com/CompVis/latent-diffusion} and the DP-LDMs paper's code\footnote{https://github.com/SaiyueLyu/DP-LDM}.
GPU Devices Used: RTX 4090, RTX 3090.

We list all the detailed parameter settings in Table~\ref{tab:autoencoder} for auto-encoder pretrain.

\begin{table}[!htb]
\tablestyle{5pt}{1.2}
\begin{tabular}{lccccc}
\toprule
Target                & CelebA-32   & CelebA-64       & MNIST           & Fashion-MNIST       & CIFAR-10    \\ \midrule
pretrain-dataset      & ImageNet   & ImageNet       & EMNIST(Letters) & CIFAR-10 (Gray scale) & ImageNet   \\
Input size            & 32         & 64             & 32              & 32                  & 32         \\
Latent size           & 16         & 32             & 4               & 4                   & 16         \\
f                     & 2          & 2              & 8               & 8                   & 2          \\
z-shape               & 16$\times$16$\times$3        & 16$\times$16$\times$3        & 4$\times$4$\times$3  &     4$\times$4$\times$3  & 32$\times$32$\times$3        \\
Channels              & 128        & 192            & 128             & 128                 & 128        \\
Channel multiplier    & {[}1,2{]}  & {[}1,2{]}      & {[}1,2,3,5{]}   & {[}1,2,3,5{]}       & {[}1,2{]}  \\
Attention resolutions & {[}16,8{]} & {[}16,8{]}     & {[}32,16,8{]}   & {[}32,16,8{]}       & {[}16,8{]} \\
Batch size            & 16         & 16             & 50              & 50                  & 16         \\
Epochs                & 4          & 10             & 50              & 50                  & 4 \\
\bottomrule
\end{tabular}%
\vspace{1pt}
\caption{Parameter settings for pretraining autoencoders.}
\label{tab:autoencoder}
\end{table}

We list all the detailed parameter settings in Table~\ref{tab:dm} for pretraining latent diffusion models.
\begin{table}[!htb]
\tablestyle{5pt}{1.2}
\begin{tabular}{lccccc}
\toprule
Targte & CelebA-32    & CelebA-64    & MNIST           & Fashion-MNIST       & CIFAR-10       \\ \midrule
pretrain-dataset           & ImageNet    & ImageNet    & EMNIST(Letters) & CIFAR-10 (Gray scale) & ImageNet      \\
model channels             & 192         & 192         & 64              & 64                  & 128           \\
channel multiplier         & {[}1,2,4{]} & {[}1,2,4{]} & {[}1,2{]}       & {[}1,2{]}           & {[}1,2,2,4{]} \\
attention resolutions      & {[}1,2,4{]} & {[}1,2,4{]} & {[}1,2{]}       & {[}1,2{]}           & {[}1,2,4{]}   \\
num res blocks             & 2           & 2           & 1               & 1                   & 2             \\
num heads                  & -           & 8           & 2               & 2                   & 8             \\
num head channels          & 32          & -           & -               & -                   & -             \\
Batch size                 & 384         & 256         & 512             & 512                 & 512           \\
Epochs                     & 40          & 40          & 120             & 120                 & 40            \\
use spatial transformer    & False       & False       & True            & True                & True          \\
cond stage key             & -           & -           & class label     & class label         & class label   \\
conditioning key           & -           & -           & crossattn       & crossattn           & crossattn     \\
num classes                & -           & -           & 26              & 10                  & 1000          \\
embedding dim              & -           & -           & 5               & 5                   & 512           \\
transformer depth          & -           & -           & 1               & 1                   & 1             \\
\bottomrule
\end{tabular}%
\vspace{1pt}
\caption{Parameter settings for pretraining latent diffusion models.}
\label{tab:dm}
\end{table}

\section{Complete Results for Unconditional Generation}
\label{sec:complete_results}
Table~\ref{tab:full_tab_1} and~\ref{tb:unconditional_generation_fid_full} show the complete results on unconditional generation on CelebA across different resolutions. 

\begin{table}[!ht]
\tablestyle{5pt}{1.2}
\begin{tabular}{l ccc}
\toprule
Method                                                 & $\epsilon=10$ & $\epsilon=5$ & $\epsilon=1$ \\ \midrule
DP-LDM (average case)                       & 14.3$\mypm$0.1 & 16.1$\mypm$0.2       & 21.1$\mypm$0.2 \\
DP-LDM (best case)                              & 14.2     & 15.8           & 21.0     \\
DP-MEPF                                           & 17.4     & 16.5           & 20.4     \\
DP-Promise                                       & 25.3     & 26.2           & 29.1     \\
PRIVIMAGE                                   & 49.3     & 52.9           & 71.4     \\
\rowcolor{Gray}
Ours (r=8, k=4, n=10,000, epoch=5)  & 14.8621  & 17.2584        & 21.4400  \\
\rowcolor{Gray}
Ours (r=8, k=4, n=60,000, epoch=5)   & 14.0125  & 16.3800        & 20.1930  \\
\rowcolor{Gray}
Ours (r=8, k=4, n=10,000, epoch=15)  & /        & /              & 16.7637  \\
\rowcolor{Gray}
Ours (r=16, k=4, n=10,000, epoch=15) & /        & /              & 15.8495  \\
\rowcolor{Gray}
Ours (r=16, k=4, n=60,000, epoch=15) & /        & /              & 15.5615  \\
\rowcolor{Gray}
Ours (r=16, k=4, n=60,000, epoch=40) & 11.2422  & 11.3459 &     14.2692     \\
\rowcolor{Gray}
Ours (r=16, k=4, n=60,000, epoch=40, project=True) & \textbf{8.4098}                     & \textbf{9.5134}                    & \textbf{12.0592}                \\ \bottomrule
\end{tabular}%
\vspace{1pt}
\caption{Ablation study with respect to different number of training samples and pre-training epochs.}
\label{tab:full_tab_1}
\end{table}

\begin{table}[!ht]
    \tablestyle{4pt}{1.2}
    \centering
    \begin{tabular}{l c c c c c c c c c}
    \toprule
        \multirow{2}{*}{Algorithm}
                                 & \multicolumn{3}{c}{CelebA-32}               &\multicolumn{3}{c}{CelebA-64}                & \multicolumn{3}{c}{CelebA-HQ} \\ 
                                 & $\epsilon=1$ & $\epsilon=5$ & $\epsilon=10$ & $\epsilon=1$ & $\epsilon=5$ & $\epsilon=10$ & $\epsilon=1$ & $\epsilon=5$ & $\epsilon=10$ \\ 
        \midrule
        \rowcolor{Gray}
        \textbf{Ours (k=4)}            & \underline{12.5} & \underline{11.9} & \underline{7.7} & \textbf{12.0}    & \textbf{9.5}     & \textbf{8.4}     & 17.2 & 18.2 & 20.0 \\
        \rowcolor{Gray}
        \textbf{Ours (k=8)} & - & - & 7.3 & - & - & 8.9 & - & - & - \\
        DP-LDMs                  & 25.8             & 16.8             & 16.2            & 21.1             & 16.1             & 14.3             &              &              &      \\
        DP-MEPF($\phi_1,\phi_2$) & 19.0             & 17.5             & 17.4            & 19.0             & 19.1             & 18.5             &              &              &      \\
        DP-MEPF($\phi_1$)        & 17.2             & 16.9             & 16.3            & \underline{18.4} & \underline{16.5} & \underline{17.4} &              &              &      \\
        PPRIVIMAGE+G             & 31.8             & 19.8             & 18.9            & 45.1             & 45.2             & 38.2             &              &              &      \\
        PPRIVIMAGE+D             & 26.0             & 20.1             & 19.1            & 71.4             & 52.9             & 49.3             &              &              &      \\
        dp-promise               & \textbf{9.0}     & \textbf{6.5}     & \textbf{6.0}    & 29.1             & 26.2             & 25.3             &              &              &      \\
        \bottomrule
    \end{tabular}
    \vspace{1pt}
    \caption{FID with unconditioned generations on CelebA across different image resolutions.}
    \label{tb:unconditional_generation_fid_full}
\end{table}

\begin{figure}[!h]
\centering
\includegraphics[width=\linewidth]{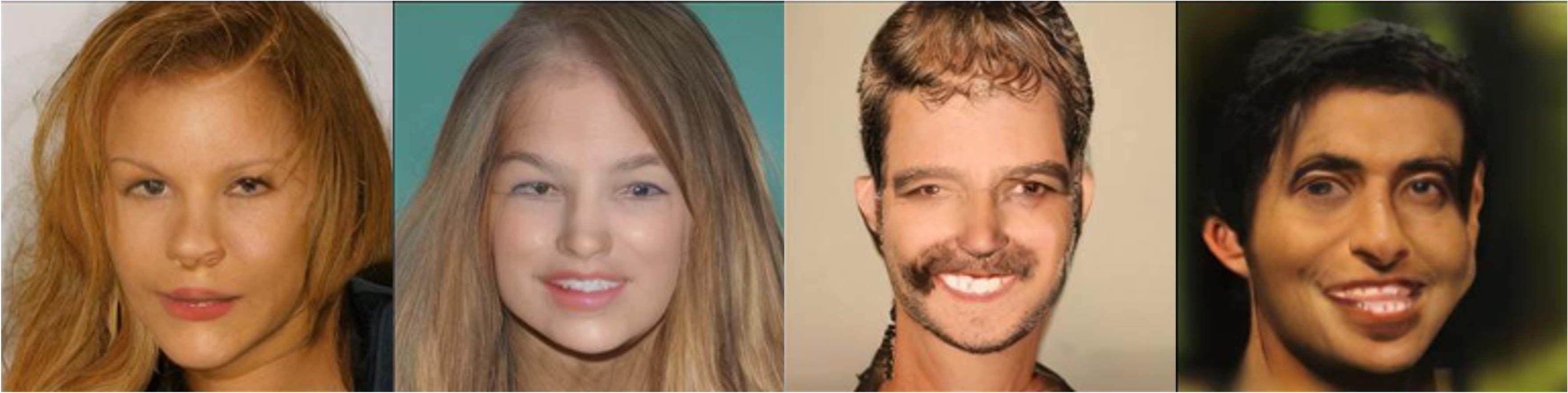}
\caption{Examples generated from CelebA-HQ with $\epsilon=10$. Because of the highly imbalance between female and male images (the number of female images is way more than the male images), the generation quality of female images are generally better than male images.}
\label{fig:hq-1}
\end{figure}

\begin{figure}[!h]
\centering
\includegraphics[width=\linewidth]{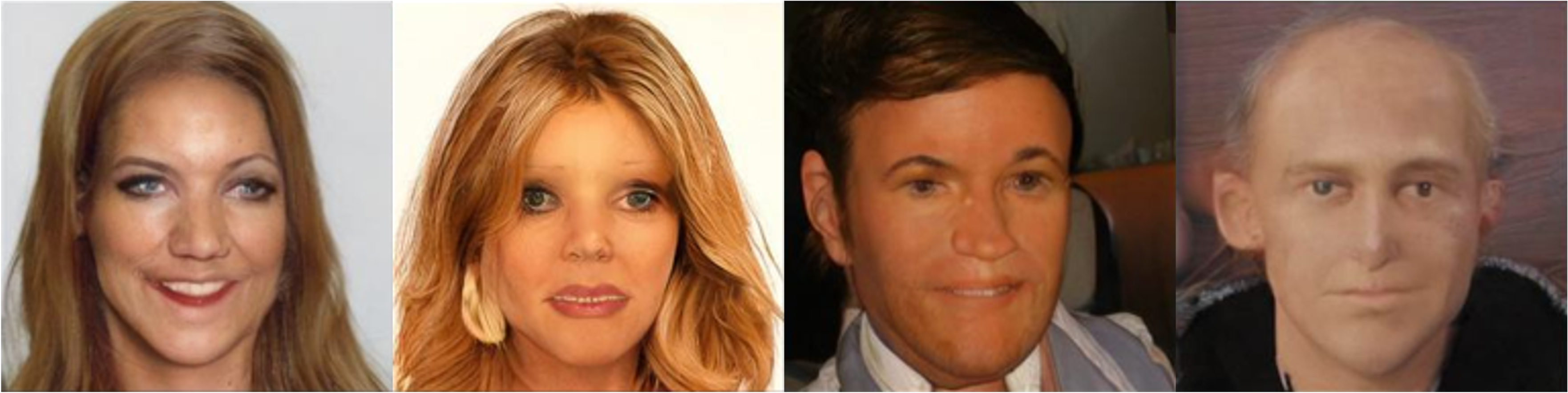}
\caption{Examples generated from CelebA-HQ with $\epsilon=5$.}
\label{fig:hq-2}
\end{figure}

\begin{figure}[!h]
\centering
\includegraphics[width=\linewidth]{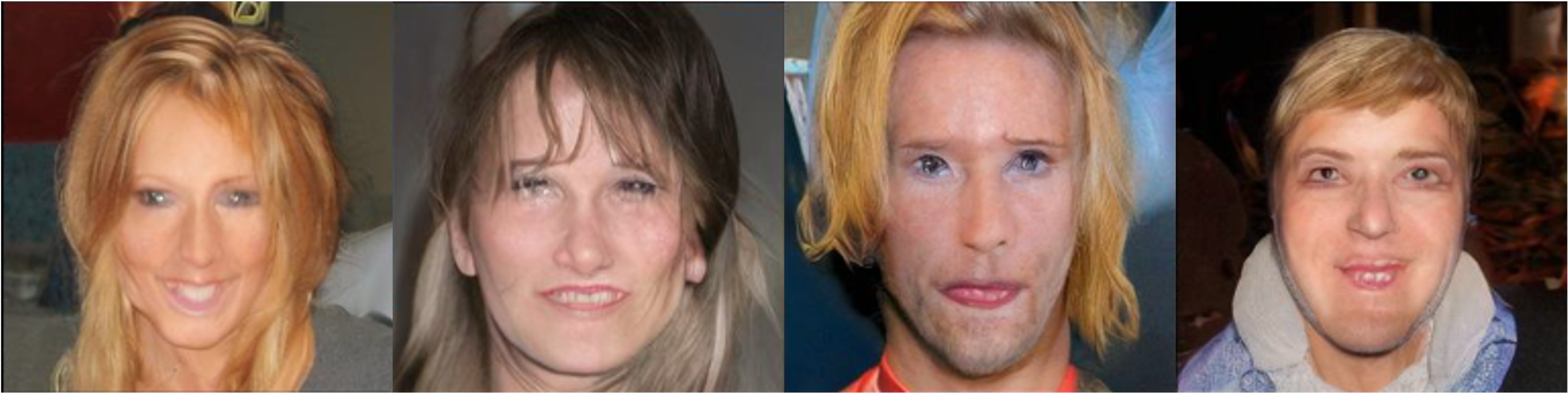}
\caption{Examples generated from CelebA-HQ with $\epsilon=1$.}
\label{fig:hq-3}
\end{figure}

Interestingly, we observe that the generation quality on female images is much better than the generation quality of male images (see Figure~\ref{fig:hq-1}~\ref{fig:hq-2}, and~\ref{fig:hq-3} which consistently across different $\epsilon$). 

\end{document}